\documentclass[runningheads]{llncs}
\usepackage[misc,geometry]{ifsym}
\usepackage{comment}
\usepackage{graphicx}
\usepackage{amsmath}
\usepackage{amssymb}
\usepackage{changepage}
\begin{document}
\title{Scene Text Recognition with Single-Point Decoding Network}

\def\YOFOSubNumber{86}  

\titlerunning{Scene Text Recognition with Single-Point Decoding Network}
%
\author{
    Lei Chen\inst{}\orcidID{0000-0001-7923-2602} \and
    Haibo Qin\inst{}\orcidID{0000-0003-2663-8830} \and
    Shi-Xue Zhang\inst{}\orcidID{0000-0001-7030-1974} \and
    Chun Yang\inst{}(\Letter)\orcidID{0000-0002-6297-4500} \and
    Xucheng Yin\inst{}\orcidID{0000-0003-0023-0220}
}
\authorrunning{L. Chen et al.}
%
\institute{University of Science and Technology Beijing, Beijing, China\\
\email{chenleiustb@163.com, hapoqin@gmail.com, zhangshixue111@163.com, \{chunyang, xuchengyin\}@ustb.edu.cn}}
\maketitle              

\begin{abstract}
In recent years, attention-based scene text recognition methods have been very popular and attracted the interest of many researchers. Attention-based methods can adaptively focus attention on a small area or even single point during decoding, in which the attention matrix is nearly one-hot distribution. Furthermore, the whole feature maps will be weighted and summed by all attention matrices during inference, causing huge redundant computations. In this paper, we propose an efficient attention-free Single-Point Decoding Network (dubbed SPDN) for scene text recognition, which can replace the traditional attention-based decoding network. Specifically, we propose Single-Point Sampling Module (SPSM) to efficiently sample one key point on the feature map for decoding one character. In this way, our method can not only precisely locate the key point of each character but also remove redundant computations. Based on SPSM, we design an efficient and novel single-point decoding network to replace the attention-based decoding network. Extensive experiments on publicly available benchmarks verify that our SPDN can greatly improve decoding efficiency without sacrificing performance.

\end{abstract}
%
%
%

\section{Introduction}
Scene text widely appears on streets, billboards, and product packaging, which contains abundant valuable information. Recently, scene text recognition has drawn much attention from researchers and practitioners, because it can be used in various applications, such as image search, intelligent inspection, and visual question answering. Although optical character recognition in scanned documents is well-developed, scene text recognition is still a challenging task due to complexity of background, diversity, variability of text, irregular arrangement, lighting and occlusion.

The development of scene text recognition can be divided into three periods. In the first period, the text recognition methods are mainly based on character segmentation \cite{ref9_wang2010word}, handcrafted features \cite{ref10_neumann2012real}, or automatic features \cite{ref15_jaderberg2014deep}.
\begin{figure}[ht]
    \centering
    \includegraphics[width=0.95\linewidth]{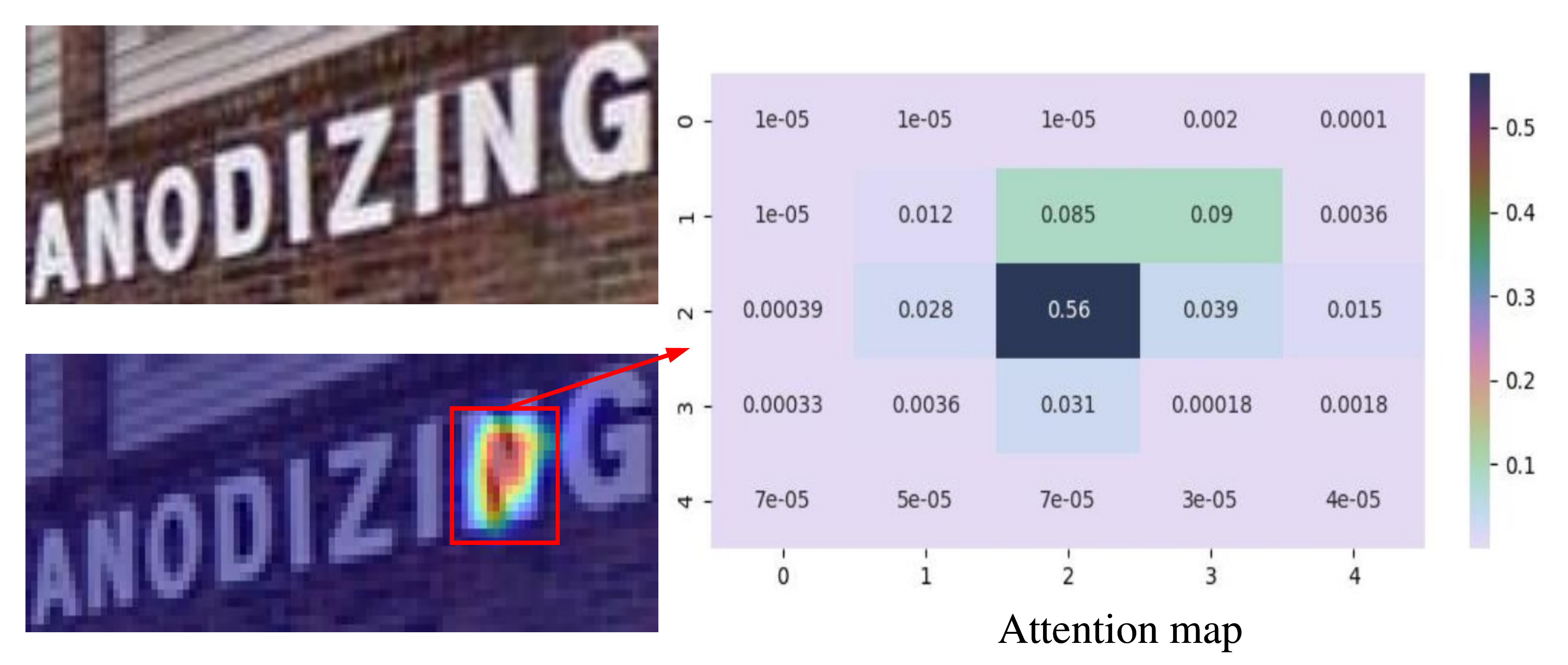}
    \caption{Demonstration of the attention map when decoding the eighth character "N". The weights of the attention map just focus on a very small area or even one point.}
    \label{fig:introduction}
\end{figure}
However, the performance of character segmentation suffers from various factors in natural scenes, such as complex background, various text styles, and irregular arrangement. In the second period, the connectionist temporal classification (CTC) based methods \cite{ref17_shi2016end} regard text recognition as sequence recognition and use the CTC algorithm \cite{ref22_graves2006connectionist} to predict symbols. However, CTC-based methods still have some limitations. One problem is that the computational cost of long text is very huge due to the complexity of the underlying implementation of the CTC algorithm. The other is that it is not easy for CTC-based methods to solve the two-dimensional text recognition problems. In the third period, benefiting from attention mechanism \cite{ref23_bahdanau2014neural}, attention-based methods have been widely popularized and achieved excellent performance.

Attention mechanism has been applied in various tasks benefiting from its excellent performance in sequence model, which can model the relationship between any two objects without considering their distance in the input or output sequences \cite{ref23_bahdanau2014neural,ref25_kim2017structured}. However, there are still some intrinsic defects in existing attention-based methods. In IFA \cite{ref55_wang2021implicit}, Wang \textit{et al}. have mathematically proved that the attention map nearly tends to a one-hot distribution in the attention mechanism. As shown in Fig. \ref{fig:introduction}, we can find that the weights of the attention map just focus on a very small area or even one point, which are much larger than the weights of the unfocused area. Obviously, the acquisition of attention maps will bring unnecessary computational cost and storage cost.

To address the above problems, in this paper, we propose an innovative and efficient \textbf{S}ingle-\textbf{P}oint \textbf{D}ecoding \textbf{N}etwork for scene text recognition, named \textbf{SPDN}, which adopts an efficient decoding network with a single point to perform decoding instead of a complex decoding network based on attention from traditional scene text recognition methods. Specifically, we proposed a \textbf{S}ingle-\textbf{P}oint \textbf{S}ampling \textbf{M}odule (\textbf{SPSM}), which can directly locate and sample one key point on the feature map for decoding one character. A position constraint loss, which is based on the priori information that the difference in character width is small, is used to improve the accuracy of sampling point positions. In this way, our method can not only locate the key point of each character but also remove redundant computation during decoding and greatly improve the decoding efficiency. Based on the proposed SPSM, we design a single-point decoding network, which can greatly improve the decoding efficiency and achieve comparable decoding accuracy against the attention-based decoding network.

In summary, the main contributions of this paper are three-fold:

1) We propose a novel Single-Point Decoding Network for scene text recognition, which adopts an efficient decoding network with a single point to perform decoding.

2) We propose a Single-Point Sampling Module (SPSM), which can directly locate and sample one key point on the feature map for decoding one character.

3) Extensive experiments on public benchmarks demonstrate that our method can not only improve decoding efficiency but also achieve promising performance.

\section{Related Work}

Deep learning technology is widely used in many fields \cite{ref67_zhang2020deep,ref68_zhang2021adaptive,ref69_hou2020detecting,ref70_zhu2020attention,ref71_zhu2021video,ref72_zhu2019residual,ref73_zhang2022arbitrary,ref74_Zhang2022KernelPN,ref75_Zhang2022GraphFN,ref76_Zhang2022ArbitrarySTN}. As a hot topic, scene text recognition is mainly divided into segmentation-based methods and sequence-based methods, and the current mainstream is the latter. The sequence-based methods predict the entire text directly from the original image without processing a single character. In sequence-based methods, the encoder-decoder framework is adopted widely. According to the structure of decoders, the sequence-based methods can be roughly divided into two categories: Connectionist Temporal Classification based (CTC-based) methods and Attention-based methods. 

\textbf{CTC-based methods} \cite{ref17_shi2016end} adopt CTC loss to train the model without any alignment between the input sequence and the output sequence. Methods of this type usually adopt convolutional neural network (CNN) to extract visual features and employ recurrent neural network (RNN) to model the contextual information. However, CTC has a large amount of computation on the long text and is difficult to be applied to two-dimensional prediction.

\textbf{Attention-based methods} \cite{ref20_lee2016recursive,ref21_cheng2017focusing} employ attention mechanism to learn the mapping between sequence features and target text sequence, which has achieved great success in text recognition. Lee \textit{et al}. \cite{ref20_lee2016recursive} used recursive CNN to obtain long contextual information, and adopted an attention-based decoder to generate target sequence. To solve the problem of irregular text recognition, Shi \textit{et al}. \cite{ref30_shi2018aster} proposed a spatial transformation network (STN) \cite{ref31_jaderberg2015spatial} to rectify the irregular text into horizontal text, and then recognized the optical character sequences in them by an attention-based recognition network. Lee \textit{et al}. \cite{ref36_lee2020recognizing} proposed the self-attention mechanism to describe the 2D spatial dependence of characters for recognizing irregular text. However, the accuracy of these methods based on the inherent decoding mechanism with attention suffers from attention drift. To solve this problem, Cheng \textit{et al}. \cite{ref21_cheng2017focusing} proposed Focus Attention Network (FAN), consisting of an attention network (AN) and a focusing network (FN), in which FN is used to adjust attention by evaluating whether AN can accurately focus on the target area. Despite their popularity and excellent performance, attention-based methods still suffer from redundant computations. In this paper, we propose a single-point decoding network to reduce computation while maintaining performance.

\section{Method}
\begin{figure*}[ht]
    \centering
    \includegraphics[width=1\textwidth]{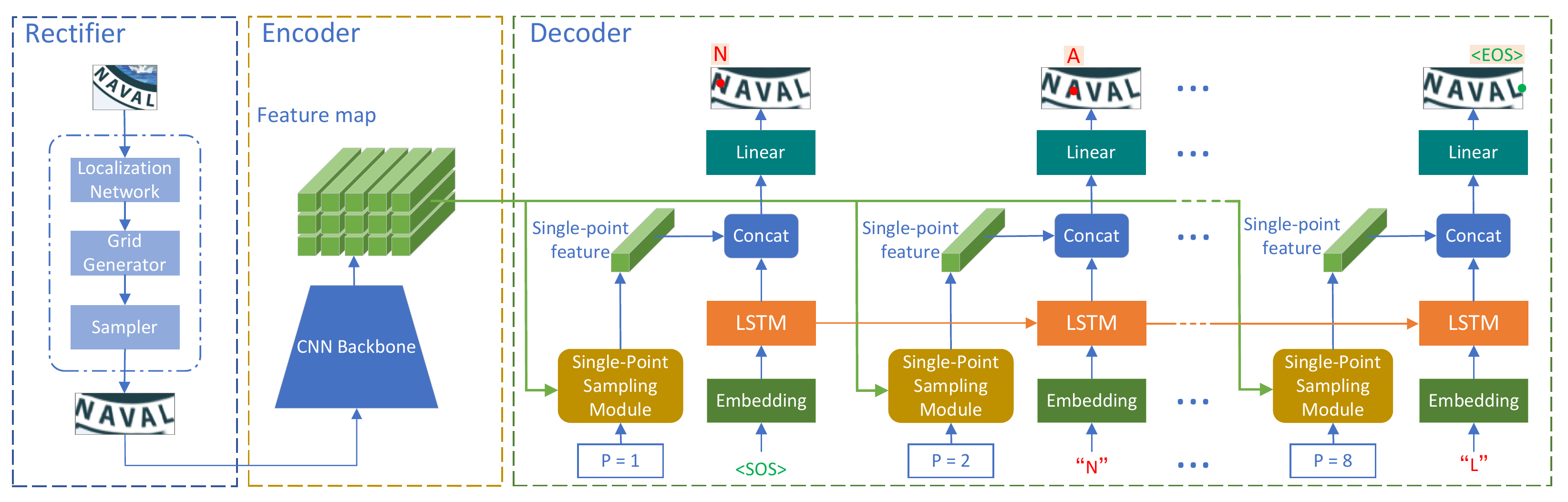}
    \caption{The overall framework of our SPDN. The pipeline of our method includes three components: a rectifier for rectifying irregular text, an encoder for extracting visual features from the rectified images, and a single-point decoder for decoding recurrently.}
    \label{fig:pipeline}
\end{figure*}

As shown in Fig. \ref{fig:pipeline}, The SPDN is composed of three parts, a rectifier for rectifying irregular text, an encoder for extracting visual features from the rectified images, and a single-point decoder for decoding recurrently and generating the final output. The rectifier consists of the localization network, the grid generator, and the sampler. After the input image is rectified by the rectification network, the encoder extracts the feature from the rectified image. Then the decoder uses LSTM to recurrently decode according to the feature. In the decoder, a single-point sampling module is designed to sample the feature of a single point on the feature map based on the priori knowledge.

\subsection{Rectifier}
Due to the superior performance of Thin Plate Spline (TPS) \cite{ref47_warps1989thin} in processing curved and perspective text, we use the STN with TPS as the framework of the rectification network. The rectification network, which is generally similar to the STN of ASTER \cite{ref30_shi2018aster}, aims to adaptively rectify irregular text from the input image and convert it into a new image.

\subsection{Encoder}
In main text recognition methods  \cite{ref49_shi2016robust,ref30_shi2018aster}, the text recognition is regarded as the task of sequence recognition, where the visual features will be converted into sequence features. However, some important information will be lost when the encoder transforms two-dimensional visual features into one-dimensional sequence features. To solve this problem, our encoder extracts the two-dimensional features and directly feeds the two-dimensional features into the decoder. The single-point sample module in our decoder can enhance the encoder’s capacity to extract discriminative features. Our decoder only uses the single point in the feature map to decode one character, which will force the feature corresponding to each character must be sufficiently concentrated at a single point.

As shown in Fig. \ref{fig:pipeline}, our encoder uses a stack of convolutional layers with residual connects as backbone to extract features. To facilitate the decoder to sample a single point on the wide two-dimensional feature maps, we will keep the feature maps of the input decoder to two-dimensional. Finally, for origin input image $\boldsymbol{x}$, we have:
\begin{equation}
    \boldsymbol{F}=\mathcal{F}(\mathcal{R}(\boldsymbol{x})), \boldsymbol{F} \in \mathbb{R}^{C\times\frac{H}{8}\times\frac{W}{8}},
\end{equation}
where $\mathcal{R}$ and $\mathcal{F}$ denote the rectifier and the encoder respectively; $C$ means the channels of the feature; $H$ and $W$ denote the height and width of the rectified image respectively.

\begin{figure}[ht]
    \centering
    \includegraphics[width=0.5\linewidth]{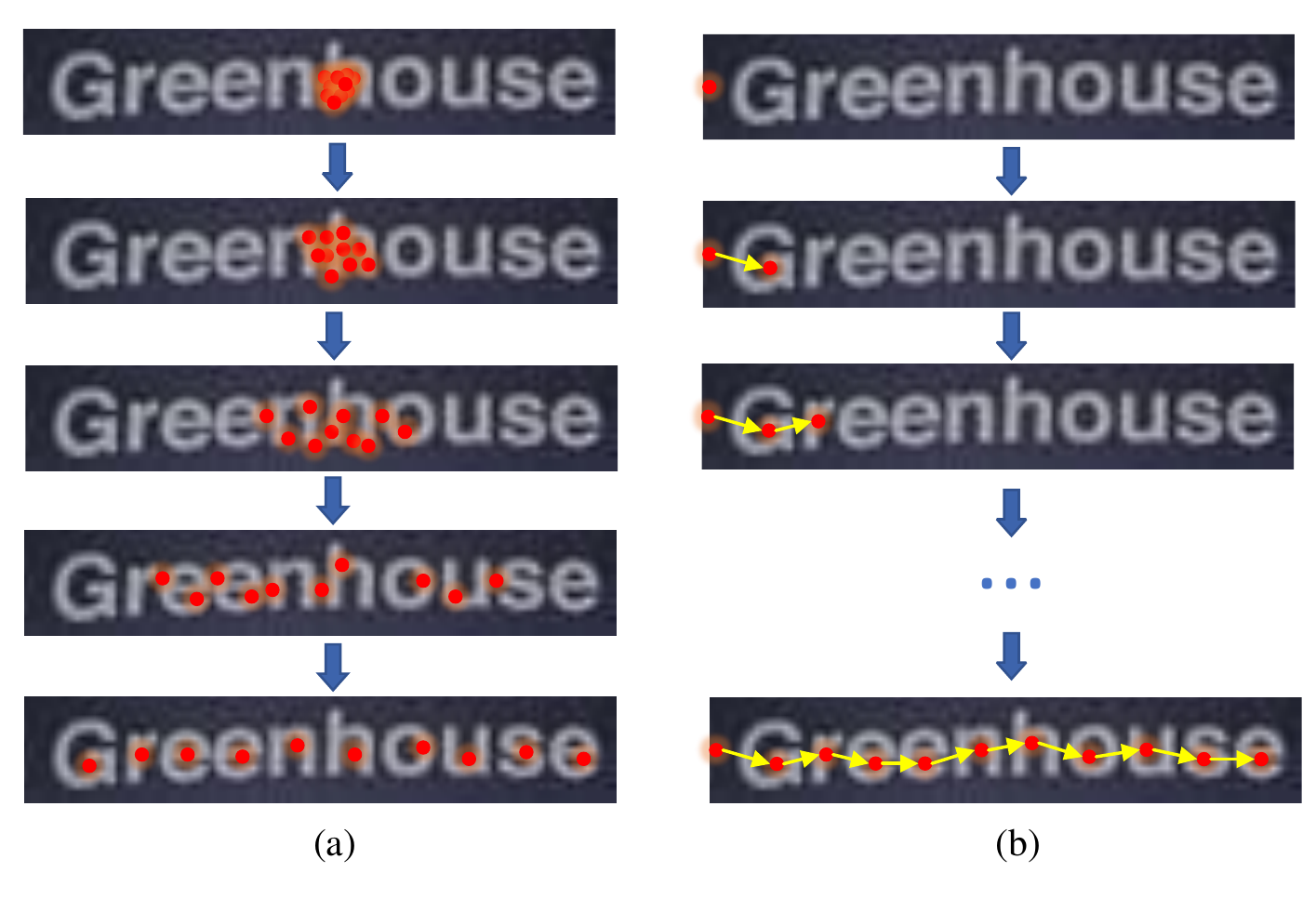}
    \caption{(a) Parallel sampling strategy. (b) Serial sampling strategy.}
    \label{fig:two_sampling_strategies}
\end{figure}

\subsection{Decoder}

As shown in Fig. \ref{fig:pipeline}, our singe-point decoding network adopts a recurrent structure with LSTM as the decoding unit and a single-point sampling module as the core role. The LSTM unit takes the hidden state at the previous moment and the embedding of the last output character as input, and generates output. As shown in Fig. \ref{fig:two_sampling_strategies}, We design two strategies for sampling the single-point feature. The parallel sampling strategy generates $T$ points initially, in which each point corresponds to a symbol. These points rely on weak constraints to learn to align the symbols, such as position embedding. The serial sampling strategy generates sampling points step by step under distance loss. From the ablation experiments in Section 4.4, we verify that the serial sampling strategy has higher performance, thus we adopt the serial sampling strategy as our sampling method.

\subsubsection{Single-Point Mechanism}
In this subsection, we will introduce the general attention mechanism and theoretically analyze its essence and elaborate on the principle of our single-point mechanism.

Usually, an attention sequence-to-sequence model is a unidirectional recurrent network which iteratively works $\boldsymbol{T}$ steps to produce $\boldsymbol{T}$ symbols including all characters in the dictionary and an end-of-sequence symbol (EOS), denoted by $(\boldsymbol{y_1},\boldsymbol{y_1},...,\boldsymbol{y_T})$. At step t, we first calculate the attention weight map through the softmax function, as shown in the following two formulas:
\begin{equation}
    e_{t,i}=\mathbf{W}_{e}\tanh(\mathbf{W}_{s}\mathbf{s}_{t-1}+\mathbf{W}_{h}\mathbf{h}_{i}+b),
\end{equation}
\begin{equation}
    \alpha_{t,i}=\frac{\exp(e_{t, i})}{\sum_{j=1}^{T}\exp(e_{t,j})},
\end{equation}
where $\mathbf{W}_{e},\mathbf{W}_{s},\mathbf{W}_{h}$ are trainable weights. $\mathbf{s}_{t-1}$ is the hidden state in the LSTM decoder at step t-1 and $\mathbf{h}_{i}$ is the $i_{th}$ vector of the encoder output $\mathbf{H}$. Then the decoder takes the attention weight map as the coefficients to calculate the context vector:
\begin{equation}
    \mathbf{c}_{t}=\sum_{i=1}^{T}\alpha_{t, i}\mathbf{h}_{i}.
\end{equation}
The context vector is fed into the recurrent cell of the decoder to generate an output vector and a new hidden state:
\begin{equation}
    \mathbf{x}_{t}, \mathbf{s}_{t}=LSTM(\mathbf{s}_{t-1},concat(\mathbf{c}_{t}, emb(y_{t-1}))),
\end{equation}
where $y_{t-1}$ is the ${t-1}$ step output symbol and $concat(\mathbf{c}_{t}, emb(y_{t-1}))$ is the concatenation of $c_t$ and the embedding of $y_{t-1}$. At last, $\mathbf{x}_{t}$ is taken to the last classifier $\mathbb{C}$ for predicting the $t$ step symbol:
\begin{equation}
    y_{t}=\mathbb{C}(\mathbf{x}_{t})=\mathbf{W}_{o}\mathbf{x}_{t}+b_{o}.
\end{equation}
According to the formulas above, we simplify the expression of $y_t$ to:
\begin{equation}
    y_t=\mathbb{C}(RNN(\sum_{i=1}^{T}\alpha_{t, i}\mathbf{h}_{i}))=\mathbf{W}\sum_{i=1}^{T}\alpha_{t, i} \mathbf{h}_{i},
\end{equation}
where $\mathbf{W}$ is the learnable weights. Since the attention map is nearly one-hot distribution, we have:
\begin{equation}
    y_t=\mathbf{W}\sum_{i=1}^{T}\alpha_{t,i}\mathbf{h}_{i} \approx \mathbf{W}\alpha_{t,i^{\prime}}\mathbf{h}_{i^{\prime}} \approx \mathbf{W}\mathbf{h}_{i^{\prime}},
\end{equation}
where $i^{\prime}=argmax(\alpha_{t,i})$, representing the attention center at $t$ step. Similarly, on the two-dimensional feature, we have the following expression:
\begin{equation}
    y_t=\mathbf{W}\sum_{i=1}^{h}\sum_{j=1}^{w}\alpha_{t,i,j}\mathbf{F}_{i,j} \approx \mathbf{W}\alpha_{t,i^{\prime},j^{\prime}}\mathbf{F}_{i^{\prime},j^{\prime}} \approx \mathbf{W}\mathbf{F}_{i^{\prime},j^{\prime}},
\end{equation}
where $\mathbf{F}$ denotes the two-dimensional feature map, $i^{\prime},j^{\prime}=argmax(\alpha_{t,i,j})$, representing the attention center on the two-dimensional feature map at $t$ step.

\subsubsection{Single-Point Sampling Module}
\begin{figure}[ht]
    \centering
    \includegraphics[width=0.95\linewidth]{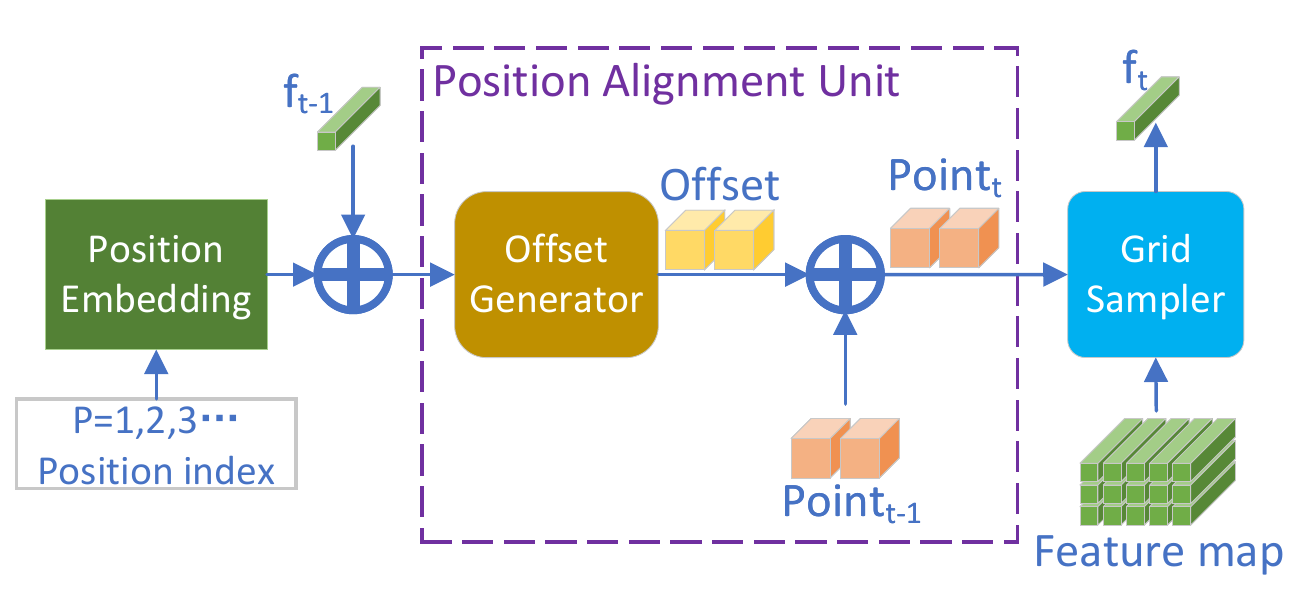}
    \caption{The illustration of the single-point sampling module.}
    \label{fig:single_point_sampling_module}
\end{figure}

As shown in Fig. \ref{fig:pipeline}, the single-point sampling module plays an important role in the decoder. The sampling module determines whether the single character feature of the current step can be accurately collected,  which will affect the performance of the decoder.

The structure of the single-point sampling module is shown in Fig. \ref{fig:single_point_sampling_module}. The single-point sampling module is composed of a position embedding layer, a position alignment unit (PAU), and a grid sampler. The character position index will be encoded by the position embedding layer. The embedding vector of the position index will be concatenated with the single-point feature of the previous moment into the PAU. The PAU contains an offset generator for generating the offset of the current sampling point from the previous sampling point. The offset generator consists of some linear layers.

\subsection{Loss Function}
In order to sample a single point more accurately, we calculate the $L_1$ loss between the current sampling point and the previous sampling point. The final objective function of the proposed method can be formulated as:
\begin{equation}
    \mathcal{L}=\mathcal{L}_{rec}+\lambda\mathcal{L}_{dist},
\end{equation}
where $\mathcal{L}_{rec}$ is the cross entropy loss between predicted symbols and groundtruth, $\mathcal{L}_{dist}$ is the $L_1$ loss between two adjacent sample points, and $\lambda$ is the balanced factor. Empirically, we set $\lambda$ to 1.

\section{Experiments}
\subsection{Datasets}
Our SPDN is trained on two publicly available synthetic datasets (Synth90K \cite{ref37_jaderberg2014synthetic} and SynthText \cite{ref38_gupta2016synthetic}) without finetuning on other datasets. We evaluate the recognition performance on six standard benchmarks, including three regular text datasets, \textit{e.g.}, ICDAR 2015 (IC15) \cite{ref41_karatzas2015icdar}, Street View Text (SVT) \cite{ref43_wang2011end}, ICDAR 2013 (IC13) \cite{ref40_karatzas2013icdar}, and three irregular text datasets, \textit{e.g.}, ICDAR 2015 (IC15) \cite{ref41_karatzas2015icdar}, Street View Text Perspective (SVTP) \cite{ref44_phan2013recognizing}, and CUTE80 (CUTE) \cite{ref45_risnumawan2014robust}.

\subsection{Implementation details}
We adopt ADADELTA as the optimizer with a batch size of 100 to minimize the objective function. The initial learning rate is set to 1, then is adjusted to 0.1 and 0.01 at the end of the 5th and 7th epochs. Our experiments are mainly carried out on 4 NVIDIA GeForce GTX 1080Ti GPUs with 11GB memory.

\begin{table*}[]
    \centering
    \caption{Results of our SPDN and SOTA methods. The second group of rows in the table denotes the comparison of our SPDN and ASTER. Numbers in bold represent best performance.}
    \setlength{\tabcolsep}{1mm}
    \renewcommand{\arraystretch}{1.1}
    \begin{tabular}{ c|c|c|c|c|c|c|c|c }
    \hline
    Methods &IIIT5K &SVT &IC13 &IC15 &SVTP &CUTE &$\begin{array}{c}\text{FLOPs} \\ \text{(G)}\end{array}$ &$\begin{array}{c}\text{Speed} \\ \text{(ms/image)}\end{array}$\\
    \hline
    CRNN \cite{ref17_shi2016end}     &81.2   &82.7   &89.6   &-      &-      &-      &- &-\\
    RARE \cite{ref49_shi2016robust} &81.9   &81.9   &88.6   &-      &71.8   &59.2   &-&-\\
    $\mathrm{R}^{2}\mathrm{AM}$ \cite{ref20_lee2016recursive} 
                                    &78.4   &80.7   &90.0   &-      &-      &-      &-&-\\
    AON \cite{ref50_cheng2018aon}   &87.0   &82.8   &-      &68.2   &73.0   &76.8   &-&-\\
    NRTR \cite{ref66_sheng2019nrtr} &90.1   &91.5   &95.8   &-      &-      &-      &29.50 &10.52\\
    SAR \cite{ref34_li2019show}     &91.5   &84.5   &91.0   &69.2   &76.4   &83.3   &32.15 &3.43\\
    CA-FCN \cite{ref27_liao2019scene}&91.9  &86.4   &91.5   &-      &-      &79.9   &-&-\\
    DAN \cite{ref56_wang2020decoupled}&93.3 &88.4   &94.2   &71.8   &76.8   &80.6   &3.77 &1.15\\
    RobustScanner \cite{ref58_yue2020robustscanner}
                                    &95.3   &88.1   &94.8   &77.1   &79.5   &\bfseries{90.3}   &15.27 &2.38\\
    SEED \cite{ref54_qiao2020seed}  &93.8   &89.6   &92.8   &80.0   &81.4   &83.6   &1.44 &1.50\\
    SATRN \cite{ref36_lee2020recognizing}
                                    &92.8   &91.3   &94.1   &79.0   &86.5   &87.8   &38.63 &10.22\\
    MASTER \cite{ref65_lu2021master}&95.0   &90.6   &95.3   &79.4   &84.5   &87.5   &14.87 &7.91\\
    ABINet \cite{ref61_fang2021read}&\bfseries{96.3} &\bfseries{93.0} &\bfseries{97.0} &\bfseries{85.0} &\bfseries{88.5} &89.2   &5.93 &2.13\\
    \hline
    ASTER \cite{ref30_shi2018aster} &93.4   &89.5   &91.8   &76.1   &78.5   &79.5   &1.11 &1.07\\
    SPDN (ours)                     &94.1   &89.9   &91.7   &77.9   &79.8   &81.6   &\bfseries{1.04} &\bfseries{0.65}\\
    \hline
    \end{tabular}
    \label{tab:sota}
\end{table*}

\subsection{Comparisons with State-of-the-Arts}
We compare our method with the state-of-the-art (SOTA) methods on 6 benchmarks in TABLE \ref{tab:sota}. The second group of rows in the table denotes the comparison of our SPDN and ASTER. Our baseline model is based on ASTER. The encoder of our SPDN is similar to the encoder of ASTER. The decoder of our method is shown in Fig. \ref{fig:pipeline}, which contains a single-point sampling module differently. Compared with ASTER, our SPDN not only surpasses it in performance, but also have faster speed. SPDN has an improvement of 0.7\%, 0.4\%, 1.8\%, 1.3\%, and 1.1\% on IIIT5K, SVT, IC15, SVTP, and CUTE respectively, which shows that our model has a significant improvement especially on irregular datasets. There is a gap between our method and the recent SOTA methods on performance, because our method is based on a relatively lightweight backbone and only uses LSTM for semantic modeling, while the recent SOTA methods mostly adopt heavy backbone networks like ResNet with FPN, and strong semantic modeling methods such as BCN language model. However, it is worth noting that compared to all recent sota method, our model has a significant advantage in speed. Specially, our method is nearly two times faster than DAN and 16 times faster than NRTR. Notably, our SPDN is efficient and can replace traditional attention-based decoding network to greatly improve the decoding efficiency.

\subsection{Ablation Study}
\subsubsection{Single-point sampling strategy}
We conduct experiments to explore the impact of two sampling strategies on performance. The performance comparison of the two sampling strategies is listed in TABLE \ref{tab:sampling_strategy}. According to TABLE \ref{tab:sampling_strategy} we can find that serial decoding outperforms parallel decoding in performance, which is because the parallel decoding does not utilize any priori information in text sequence. Our experiments are based on serial strategies.
\begin{table}[]
    \centering
    \caption{The performance of the two sampling strategies.}
    \begin{tabular}{ |c|c|c|c|c|c|c| }
    \hline
    $\begin{array}{c}
        \text{Sampling} \\
        \text{strategy}
    \end{array}$    &IIIT5K  &SVT    &IC13   &IC15   &SVTP   &CUTE\\
    \hline
    $\begin{array}{c}
        \text{Parallel} \\
        \text{sampling}
    \end{array}$    &93.6    &87.4   &89.4   &76.2   &76.7   &79.9\\
    \hline
    $\begin{array}{c}
        \text{Serial} \\
        \text{sampling}
    \end{array}$    &94.1    &89.9   &91.7   &77.9   &79.8   &81.6\\
    \hline
    \end{tabular}
    \label{tab:sampling_strategy}
\end{table}

\begin{table}[]
    \centering
    \caption{Impact of the number of key points on performance. "\#" means the number of key points acquired by the single-point sampling module.}
    \setlength{\tabcolsep}{0.9mm}
    \begin{tabular}{ |c|c|c|c|c|c|c|c| }
    \hline
    $\begin{array}{c}
        \text{\# Key} \\
        \text{points}
    \end{array}$    &IIIT5K &SVT    &IC13   &IC15   &SVTP   &CUTE   &$\begin{array}{c}\text{Speed} \\ \text{(ms/image)}\end{array}$\\
    \hline
    1               &94.1   &89.9   &91.7   &77.9   &79.8   &81.6   &0.65\\
    \hline
    2               &94.2   &89.9   &91.8   &77.8   &79.4   &81.9   &0.74\\
    \hline
    3               &94.4   &90.1   &92.0   &78.0   &79.6   &81.8   &0.81\\
    \hline
    4               &94.5   &89.8   &92.1   &78.5   &79.7   &81.9   &0.88\\
    \hline
    \end{tabular}
    \label{tab:keypoints_number}
\end{table}

\subsubsection{Impact of the number of key points on performance}
In this subsection, we conduct experiments to explore the influence of the number of key points on performance. As shown in TABLE \ref{tab:keypoints_number}, with the increase of the number of key points, the recognition performance increases slowly, but the decoding speed decrease significantly. It is experimentally shown that the feature of each character can be fully expressed by only one key point.


\begin{figure*}[ht]
    \centering
    \includegraphics[width=0.95\linewidth]{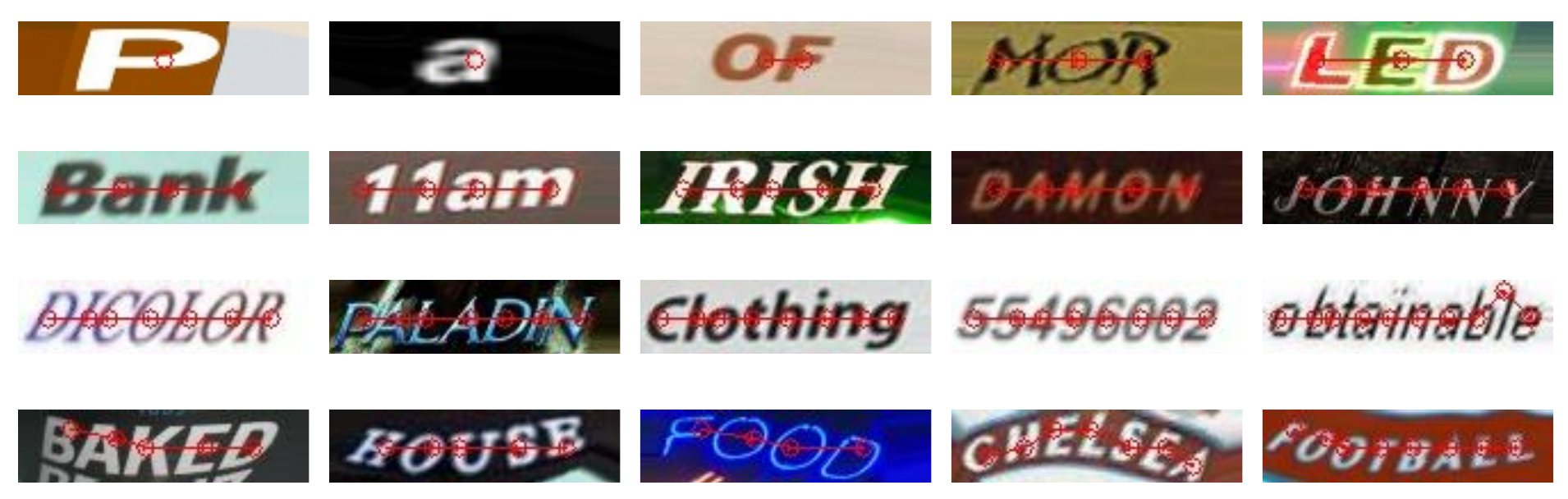}
    \caption{Visual examples of the single-point decoding process.}
    \label{fig:vis}
\end{figure*}

\subsection{Visual illustrations}
Fig. \ref{fig:vis} shows the visualization of single-point decoding, the first row shows short words that are less than four in length; the second row shows medium words of length 4-6; the third row shows long words that are more than six in length; the last row shows the curved text. According to Fig. \ref{fig:vis}, we can find that our single-point sampling module can adaptively locate the position of the key points on the text with any length and shape.

\section{Conclusion}
In this paper, we propose a novel scene text recognition network, named single-point decoding network (SPDN), which can use a single-point feature to decode one character. In addition, we propose a single-point sampling module, which can use the priori information to accurately and adaptively locate key points on the text with any length and shape. Our SPDN can greatly improve decoding efficiency and achieve comparable performance.

\section{Acknowledgment}
The research is partly supported by the National Key Research and Development Program of China (2020AAA09701), The National Science Fund for Distinguished Young Scholars (62125601), and the National Natural Science Foundation of China (62006018, 62076024).

\bibliographystyle{splncs04}
\bibliography{ref.bib}





\end{document}